\pdfoutput=1

\documentclass[11pt]{article}

\usepackage[preprint]{acl}

\usepackage{times}
\usepackage{latexsym}
\usepackage{multirow}
\usepackage{multicol}
\usepackage{algorithm}
\usepackage{amsmath}
\usepackage{booktabs}
\usepackage{amssymb}
\usepackage{enumitem}

\usepackage{algpseudocode}
\usepackage{makecell}

\usepackage[T1]{fontenc}

\usepackage[utf8]{inputenc}

\usepackage{microtype}

\usepackage{inconsolata}

\usepackage{graphicx}

\usepackage{wrapfig}

\usepackage{pgfplots}
\usepackage{xcolor}
\usepackage{float}

\usepackage{bm}%

\newcommand{\bx}{\bm{x}}

\newcommand{\sto}{\mbox{\normalfont s.t.}}
\definecolor{darkblue}{rgb}{0, 0, 0.5}
\hypersetup{colorlinks=true, citecolor=darkblue, linkcolor=darkblue, urlcolor=darkblue}

\newcommand{\StartMenu}{\raisebox{-0.18cm}{\includegraphics[scale=0.2]{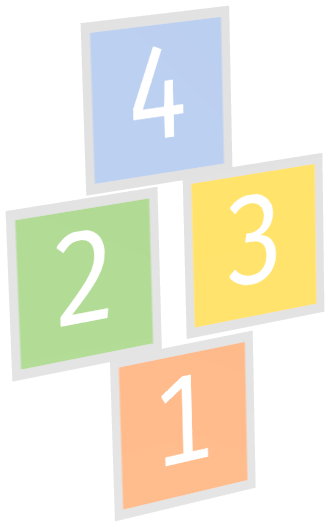}}}%

\title{\StartMenu~\texttt{Hopscotch}: Discovering and Skipping Redundancies \\in Language Models}

\author{
 \textbf{Mustafa Eyceoz\textsuperscript{1}},
 \textbf{Nikhil Shivakumar Nayak\textsuperscript{1}},
 \textbf{Hao Wang\textsuperscript{1}},
 \textbf{Ligong Han\textsuperscript{1}},
 \textbf{Akash Srivastava\textsuperscript{1}}
\\
\\
 \textsuperscript{1}Red Hat AI Innovation
\\
 \small{
   \textbf{Correspondence:} \href{mailto:meyceoz@redhat.com}{meyceoz@redhat.com}
 }
}

\begin{document}

\maketitle

\begin{abstract}
Modern causal language models stack many attention blocks to improve performance, but not all blocks are necessary for every task. We propose Hopscotch, a simple yet effective method that identifies and skips attention blocks with least contributions to a task and adapts to preserve output quality. Hopscotch jointly optimizes which blocks to skip and how to scale the outputs of the remaining layers. 
By introducing lightweight, trainable scaling parameters to attention and MLP blocks, it mitigates distribution shifts in hidden states caused by removing attention blocks. 
Hopscotch does not modify model weights or require access to pretraining or instruction-tuning data, and is compatible with existing model compression techniques. 
When applied to \texttt{Llama-3.1-8B} and \texttt{Qwen2.5-7B}, Hopscotch achieves less than a 2\% drop in performance even after skipping four attention blocks. 
\end{abstract}

\section{Introduction}

Large language models (LLMs) continue to grow in size, driven by ``scaling laws'' that suggest larger models tend to yield better performance \citep{hestness2017deep,hoffmann2022training,henighan2020scaling}. Adding more attention blocks increases model capacity, but self-attention is also the most expensive operation in LLMs. Unlike MLP blocks, attention incurs a quadratic computational cost with respect to sequence length, making it a dominant factor in inference-time efficiency. 
However, not all attention blocks are equally important for every task, and some may carry redundant information. In this paper, we explore whether entire attention blocks can be skipped without significant performance degradation.

We introduce \textbf{Hopscotch}\footnote{This name is inspired by the classic game where players hop through numbered squares, skipping the one with the marker.}, a method that jointly learns which attention blocks to skip and how to rescale the outputs of the remaining attention and MLP blocks. Hopscotch iteratively identifies attention blocks with minimal contribution to the target task and introduces lightweight, trainable scaling parameters to mitigate distribution shifts in hidden states caused by block removal. %
Hopscotch requires no changes to model weights and no access to pretraining or instruction-tuning data. %
Additionally, it is compatible with fine-grained compression techniques, such as model sparsification \citep{frantar2023sparsegpt} or KV cache quantization \citep{liu2024kivi,hooper2024kvquant,wang2025squat}, and can be combined with them to further reduce LLM inference costs.

We evaluate Hopscotch on instruction-tuned models, including {\small\texttt{Llama-3.1-8B-Instruct}} and {\small\texttt{Qwen2.5-7B-Instruct}}, and find that it can successfully skip up to 7 attention blocks with less than a 3\% average performance drop across diverse tasks while yielding up to 15\% inference speedup (see Section~\ref{app:inference}). With 4 blocks removed, we retain over 98.6\% of baseline accuracy. These results highlight structural redundancy in attention blocks. We measure distributional shift in hidden representations and find that Hopscotch significantly reduces deviation from the original model, compared to unscaled attention block skipping. Finally, we show that Hopscotch is compatible with leading quantization techniques, such as GPTQ \citep{frantar2022gptq} and AWQ \citep{lin2024awq}.

\section{Related Work}

\noindent\textbf{Feature scaling in generative models.} Our work is inspired in part by recent advances in feature modulation techniques in generative models. For example, FreeU~\citep{si2024freeu} enhances image quality in diffusion U-Nets by scaling hidden states and skip connections %
and \citet{ma2024surprising} show that channel-wise scaling during post-training inference improves diffusion sampling quality. %
Our work extends this direction by proposing dynamic rescaling of intermediate features in LLMs to compensate for pruned attention blocks.

\noindent\textbf{Skipping computation in LLMs.} Several recent studies have explored the possibility of reducing computation in LLMs by selectively skipping parts of the model. \citet{shukor2024skipping} show that in multimodal LLMs, certain attention blocks contribute less to performance in vision-language tasks and can be skipped in these cases with minimal degradation. Further investigations on the true impact of attention blocks at different depths \citep{ben2024attend, he2024matterstransformersattentionneeded} find that the effect is highly block-specific. %
Our work builds on these insights by providing a principled mechanism to identify and compensate for less informative attention blocks using learned scaling.

\noindent\textbf{Model compression.} Model compression techniques, including pruning, sparsification, and quantization, have been widely studied as means to reduce the memory and computational footprint of LLMs~\citep[e.g.,][]{frantar2023sparsegpt,sun2023simple,frantar2022gptq,xiao2023smoothquant,yao2022zeroquant,ashkboos2025quarot,chee2023quip}. These methods have shown impressive reductions in model size with minimal performance loss. %
Our approach is orthogonal to these methods, and can be used separately or in combination to further enhance model efficiency.

\section{The Hopscotch method}

\begin{figure}[t]
    \centering
  \includegraphics[width=0.99\columnwidth]{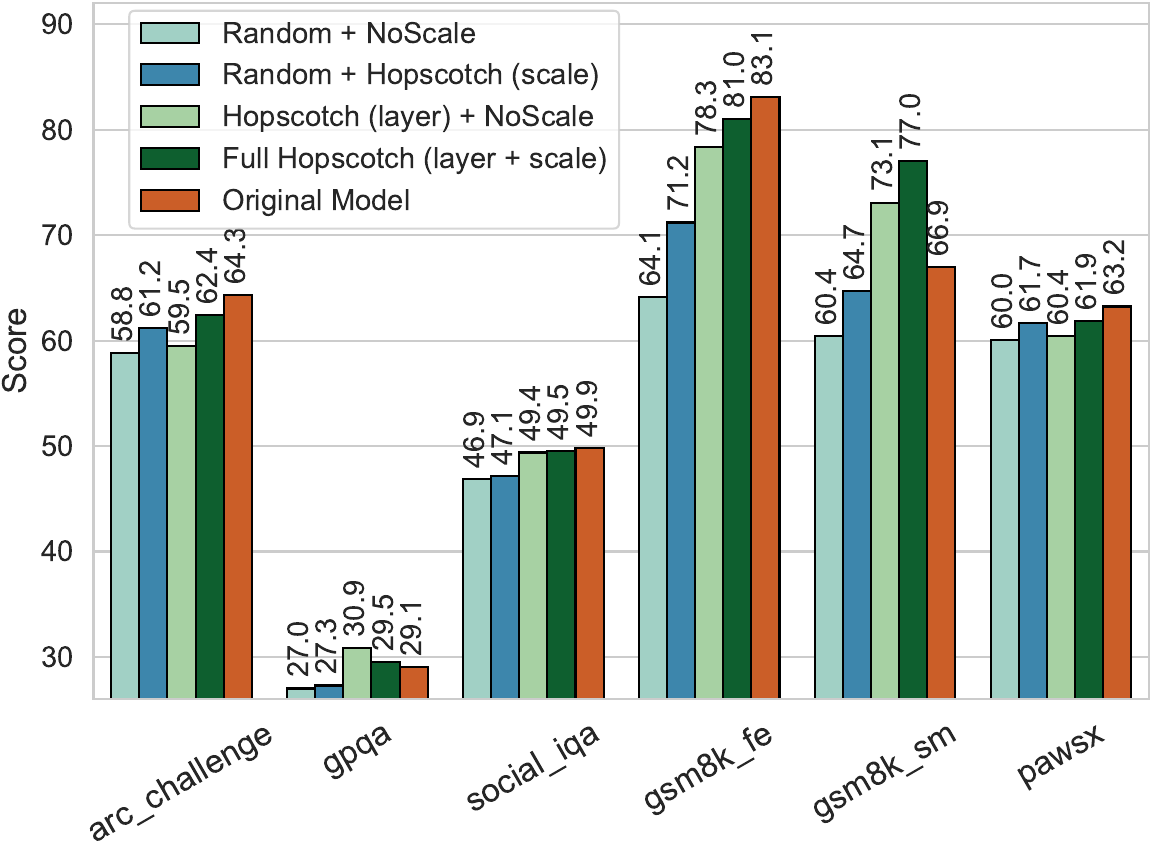}
  \caption{\footnotesize Comparison of benchmark performance when four attention blocks from layers (14, 17, 21, 24) are removed, before and after training scaling factors using Hopscotch. We also report results from Hopscotch's full pipeline: block selection and rescaling. Baseline scores from the original model without any blocks removed are included for reference.}
\label{fig:benchmark_comparison_wide_acl_final}
\end{figure}

\paragraph{Problem setup.} 
When skipping one attention block, the input distribution to the next layer changes. To compensate the change, for each layer, we introduce four trainable scalar factors: one each for the outputs of the attention block, the MLP block, and the two residual connections. Let $\bx_{\text{in}}^{(l)} \in \mathbb{R}^{n \times d}$ denote the input to the $l$-th transformer block, where $n$ is the sequence length and $d$ is the hidden dimension. %
Each transformer block is modified as follows: 
\begin{align*}
\bx_1^{(l)} &= \colorbox{cyan!30}{$b_{\mathsf{att}}^{(l)}$}  \mathsf{Attention}(\mathsf{Norm}(\bx_{\mathsf{in}}^{(l)})) + \colorbox{cyan!30}{$s_{\mathsf{att}}^{(l)}$}  \bx_{\mathsf{in}}^{(l)} \\
\bx_{\mathsf{out}}^{(l)} &= \colorbox{cyan!30}{$b_{\mathsf{mlp}}^{(l)}$} \mathsf{MLP}(\mathsf{Norm}(\bx_1^{(l)})) + \colorbox{cyan!30}{$s_{\mathsf{mlp}}^{(l)}$} \bx_1^{(l)}.
\end{align*}
Setting all scaling factors to 1.0 recovers the original model, while setting any factor to 0.0 disables (i.e., remove) the corresponding component. %

\begin{table*}[h]
    \centering
    \footnotesize 
    \setlength{\tabcolsep}{4pt} 

    \begin{tabular}{l l *{6}{>{\raggedleft\arraybackslash}p{1.6cm}}} 
    \toprule
    \multicolumn{2}{l}{\textbf{Method}} & \textbf{gsm8k FE} & \textbf{gsm8k SM} & \textbf{arc\_challenge} & \textbf{gpqa} & \textbf{social\_iqa} & \textbf{pawsx} \\
    \midrule
    \multicolumn{2}{l}{Baseline} & 83.10 & 66.94 & 64.33 & 29.07 & 49.85 & 63.26 \\
    \midrule
    \multirow{3}{*}{\textbf{4 Blocks}} 
    & Hopscotch & \textbf{81.05} & \textbf{77.03} & \textbf{62.42} & \textbf{29.53} & \textbf{49.49} & \textbf{61.90} \\
    & ShortGPT  & 62.33 & 58.45 & 57.00 & 28.44 & 47.59 & 59.49 \\
    & FinerCut  & 75.80 & 61.00 & 53.58 & 27.77 & 46.93 & 54.98 \\
    \midrule
    \multirow{3}{*}{\textbf{7 Blocks}} 
    & Hopscotch & \textbf{79.38} & \textbf{75.82} & \textbf{61.17} & \textbf{29.39} & \textbf{48.93} & \textbf{60.44} \\
    & ShortGPT  & 1.90  & 1.29  & 46.42 & 27.35 & 43.76 & 57.62 \\
    & FinerCut  & 42.50 & 6.80  & 52.65 & 28.10 & 46.47 & 54.87 \\
    \midrule
    \textbf{10 Blocks} & Hopscotch & \textbf{67.93} & \textbf{62.58} & \textbf{57.80} & \textbf{29.83} & \textbf{48.23} & \textbf{61.06} \\
    \bottomrule
\end{tabular}
    \caption{Accuracy (\%) of {\small\texttt{Llama-3.1-8B-Instruct}} on various benchmarks after skipping attention blocks (or full decoder blocks for ShortGPT). The table shows the scores of the original model compared to Hopscotch as well as prior methods when skipping 4, 7, or 10 blocks.}
    \label{tab:combined_recovery_results_final}
\end{table*}

\paragraph{Loss function.} Suppose we have an instruction-tuning dataset $\{x^{(i)}\}_{i=1}^{N}$, where each $x^{(i)}$ is a prompt consisting of an instruction and its input. We use the LLM on which we aim to skip attention blocks to generate the corresponding response $y^{(i)}$. Let $L$ be the number of transformer blocks. We define the following loss function to learn the scaling factors $\theta = \left\{b_{\mathsf{att}}^{(l)}, s_{\mathsf{att}}^{(l)}, b_{\mathsf{mlp}}^{(l)}, s_{\mathsf{mlp}}^{(l)}\right\}_{l=1}^L$:
\begin{equation*} 
\mathcal{L}\left(\theta\right) = \frac{1}{N} \sum_{i=1}^{N} \left( - \sum_{t=1}^{T_i} \log P_{\theta}(y_t^{(i)} \mid y_{<t}^{(i)}, x^{(i)}) \right).
\end{equation*}
The model weights are frozen, and only the scaling factors are updated during training. See Appendix~\ref{app:loss} for further details on loss function selection and recovery data.%

\subsection{Greedy Iterative Algorithm}

At the heart of the Hopscotch method is a greedy iterative algorithm. In each iteration, we identify an attention block whose removal minimally affects the model's output, remove it, and then rescale the remaining blocks to compensate. This process is repeated until a target number of blocks are pruned or a performance threshold is reached. Below, we provide the detailed procedures.

\paragraph{Selecting attention blocks to remove.}
To select an attention block for removal, we estimate the impact of removing each block on the model’s loss. Specifically, we define the impact of removing the $l$-th attention block as:  $\min \mathcal{L}\left(\theta\right)~\sto~b_{\mathsf{att}}^{(l)} = 0$. Solving this optimization exactly for \emph{every} layer is computationally expensive. Instead, we approximate it by running a single optimization epoch with a large learning rate, quickly estimating which layer can be removed with minimal degradation. %

\paragraph{Rescaling the remaining blocks.}
Once an attention block is removed, we rescale the remaining blocks to recover model performance. This is done by minimizing the loss function $\mathcal{L}(\theta)$ over multiple epochs using a small learning rate. This longer optimization process adjusts the scaling factors to best compensate for the removed block and recovers model quality.
Exact learning rates and hyperparameters can be found in Appendix~\ref{app:hyper}.

\section{Numerical Experiments}

\subsection{Benchmarks and Setup} 
To assess the affects of attention block skipping and the Hopscotch method, we took a sample set of benchmarks: ARC Challenge (Abstraction and Reasoning) \citep{clark2018think}, GPQA (Challenging Google-Proof QA) \citep{rein2024gpqa}, Social IQA (Commonsense Social Reasoning) \citep{sap2019socialiqa}, GSM8K (General Math) \citep{cobbe2021training}, PAWS-X (Multilingual) \citep{yang2019paws}.

\begin{figure}[t]
    \centering
  \includegraphics[width=0.865\columnwidth]{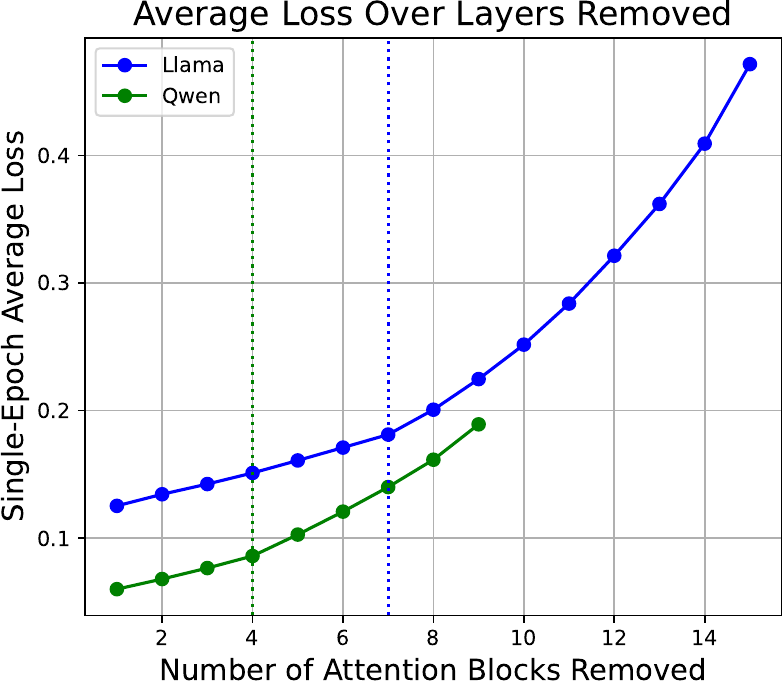}
  \caption{A plot of the single-epoch average loss per block removed during step one of Hopscotch for Llama.}
  \label{fig:removal}
\end{figure}

We run Hopscotch on {\small\texttt{Llama-3.1-8B-Instruct}} and {\small\texttt{Qwen2.5-7B-Instruct}} (see Appendix~\ref{app:model} for details on model choice). As shown in Figure~\ref{fig:removal}, during the attention block removal step, the removal of initial blocks results in relatively small, approximately linear increases in loss up to the seventh block for Llama and the fourth for Qwen. Based on this observed inflection in the loss curve, consistent with the elbow method heuristic~\citep{wu2022filter}, we use four and seven block removals as representative configurations for our evaluations.

Figure~\ref{fig:benchmark_comparison_wide_acl_final} compares the benchmark performance of our full method, Hopscotch layer selection with scaling, against four baselines: (i) the original unmodified model (baseline), (ii) random block removal with no scaling, (iii) random block removal with Hopscotch scaling, and (iv) Hopscotch-selected layers with no scaling. When simply removing sets of arbitrary attention blocks near the center of the model, performance remained within a few points for all but GSM8K, which degraded drastically. When using Hopscotch scaling, even without layer selection, we were able to recover a significant portion of the GSM8K performance, while also seeing slight improvement in other domains. For this reason, we use the math domain and GSM8K as our primary test case and training recovery set for the Hopscotch method, while continuing to evaluate on the full set of  benchmarks.

\subsection{Results on Llama-3.1-8B-Instruct}

In Table~\ref{tab:combined_recovery_results_final}, we present the results for {\small\texttt{Llama-3.1-8B}}-{\small\texttt{Instruct}} with four and seven blocks removed via Hopscotch, evaluated on the full benchmark suite. With seven blocks removed, we observe an average benchmark performance retainment of 97.08\% when ignoring the GSM8K-SM (strict match) increase, and including it we see an average of 99.78\%. With four blocks removed, average  performance in this case drops only 1.35\% with a 98.65\% overall recovery even when excluding strict match score. With strict match included average score actually goes up by 1.39\%.

We also compare directly to two leading prior works: SparseGPT \cite{frantar2023sparsegpt} and FinerCut \cite{zhang2024finercutfinergrainedinterpretablelayer}. Hopscotch consistently outperforms both, especially on challenging tasks such as GSM8K, where other methods degrade significantly after removing four blocks and fail entirely beyond seven. Against ShortGPT, a key note is that the method removes entire layers, rather than keeping MLP blocks intact. This means for a fair comparison in terms of inference-time efficiency, one should compare ShortGPT with four layers removed to Hopscotch with at least six attention blocks removed (see Section 4.6 on how attention blocks contribute to inference time). We show that even with seven blocks removed, Hopscotch still outperforms across the board. For a closer comparison in terms of parameters removed, with 10 attention blocks removed, Hopscotch corresponds to about 83.3\% of the parameter reduction by ShortGPT with 4 layers, yet achieves a relative inference speedup of \(\mathbf{165.06\%}\). In this setting, Hopscotch still outperforms ShortGPT with 4 layers across all benchmarks (see Table~\ref{tab:combined_recovery_results_final}). FinerCut also only removes attention blocks, making comparisons balanced with equal block removal. The primary remaining differences lie in the method of candidate block selection, as well as the critical discovery of the importance of scaling parameters.

\subsection{Results on Qwen-2.5-7B-Instruct}

Our findings also hold when using {\small\texttt{Qwen2.5-7B}}-{\small\texttt{Instruct}}. Table~\ref{tab:results_qwen} shows the results for four blocks removed via the Hopscotch method.
We observe an average recovery of 98.1\% in benchmark performance when excluding GSM8K-SM (strict match). On-task GSM8K performance shows essentially perfect recovery. This is not surprising, as GSM8K training data was used to learn the scaling factors in Hopscotch, though the level of recovery notably exceeds the Llama experiments.

\begin{table}[h]
    \centering
    \renewcommand{\arraystretch}{1.1} %
    \begin{tabular}{lccc}
        \toprule
        \multirow{1}{*}{\textbf{Bench}} 
        & \textbf{4 Blocks} & \textbf{Baseline} & \textbf{Recovery}\\
        \hline
        gsm8k FE & 73.16 & 73.24 & 99.89\% \\
        gsm8k SM & 39.20 & 17.74 & 220.97\% \\
        \hline
        arc\_chall & 57.51 & 59.81 & 96.15\% \\
        gpqa & 29.53 & 29.87 & 98.86\% \\
        social\_iqa & 43.81 & 45.96 & 95.32\% \\
        pawsx & 60.14 & 59.97 & 100.28\% \\
        \bottomrule
    \end{tabular}
    \caption{{\small\texttt{Qwen2.5-7B-Instruct}} performance (accuracies) after skipping 4 attention blocks via Hopscotch.}
    \label{tab:results_qwen}
\end{table}

\subsection{Measuring Distributional Shift}

Removing attention blocks introduces a shift in hidden state representations throughout the model. To quantify this shift and evaluate how Hopscotch scaling effectively mitigates it, we compute the Maximum Mean Discrepancy (MMD) between hidden states in the original {\small\texttt{LLaMA-3.1-8B-Instruct}} model and its modified variants. MMD is a non-parametric metric used to quantify the difference between two distributions based on samples drawn from them. Specifically, we compare the MMD between (i) the original model and the version with zeroed attention blocks and no rescaling (``NoScale''), and (ii) the original model and the Hopscotch-scaled model.

\begin{table}[h]
    \centering
    \resizebox{1\linewidth}{!}{
    \begin{tabular}{cccccccc}
        \toprule
        \textbf{Layer \#} & \textbf{28} & \textbf{26} & \textbf{23} & \textbf{22} & \textbf{21} & \textbf{19} & \textbf{4} \\
        \midrule
        \textbf{NoScale} & 0.62 & 0.50 & 0.40 & 0.38 & 0.33 & 0.26 & 0.35 \\
        \textbf{Ours} & \textbf{0.40} & \textbf{0.31} & \textbf{0.25} & \textbf{0.22} & \textbf{0.19} & \textbf{0.11} & \textbf{0.34} \\
        \bottomrule
    \end{tabular}}
    \caption{MMD scores across layers for two intervention methods. First row: Original vs. NoScale; Second row: Original vs. Ours.}
    \label{tab:mmd_7_layer}
\end{table}

As shown in Table~\ref{tab:mmd_7_layer}, Hopscotch consistently yields lower MMD from the original model than NoScale across all layers, indicating reduced distributional shift. For instance, in layer 19, MMD drops from 0.2598 (NoScale) to 0.1098 with Hopscotch, a reduction of over 57\%. These results support the core mechanism of Hopscotch: post-hoc scaling restores internal representations after attention block removal.

\subsection{Compatibility with Quantization}
To evaluate the compatibility of Hopscotch with other post-training model compression methods, we consider two state-of-the-art quantization techniques: GPTQ~\citep{frantar2022gptq} and AWQ~\citep{lin2024awq}. We test two strategies: (i) applying Hopscotch after quantizing the model, and (ii) applying Hopscotch before quantization. For both approaches, we report performance for 4 and 7 skipped attention blocks. As shown in Table~\ref{tab:quant_hopscotch}, Hopscotch remains effective when combined with quantization irrespective of the application order. It consistently improves strict match accuracy and achieves flexible extract performance comparable to the original uncompressed model.


\begin{table}[h]
    \centering
    \resizebox{1.0\linewidth}{!}{
    \begin{tabular}{lcc}
        \toprule
        \textbf{Method} & \textbf{Strict Match} & \textbf{Flexible Extract} \\
        \midrule
        Baseline (Instruct) & 67.0 & 83.9 \\
        \midrule
        GPTQ & 68.0 & 83.9 \\
        GPTQ + Hopscotch (4) & 77.3 & 79.4 \\
        GPTQ + Hopscotch (7) & 75.1 & 78.5 \\
        Hopscotch (4) + GPTQ & 77.9 & 79.2 \\
        Hopscotch (7) + GPTQ & 73.8 & 78.5 \\
        \midrule
        AWQ & 65.0 & 81.2 \\
        AWQ + Hopscotch (4) & 71.4 & 75.1 \\
        AWQ + Hopscotch (7) & 69.5 & 73.8 \\
        Hopscotch (4) + AWQ & 75.2 & 78.3 \\
        Hopscotch (7) + AWQ & 73.1 & 79.2 \\
        \bottomrule
    \end{tabular}
    }
    \caption{GSM8K evaluation results for {\small\texttt{LLaMA-3.1-8B}}-{\small\texttt{Instruct}} using Hopscotch in combination with GPTQ and AWQ quantization.}
    \label{tab:quant_hopscotch}
\end{table}

\subsection{Effects on Model Efficiency}
\label{app:inference}

In a given forward pass for {\small\texttt{Llama-3.1-8B-Instruct}}, we experimentally find that $\sim$66\% of time is spent in decoder attention blocks. In a model with 32 hidden layers, this means $\sim$2.06\% of the forward pass is spent in each block. For example, for a query with an average forward pass time of 0.054 seconds, the time spent in a single attention block is on average 0.0011 seconds. The effect of removing a subset of attention blocks on overall inference time is summarized in Table~\ref{tab:inference_gain}.

\begin{table}[h]
    \centering
    \begin{tabular}{cc}
        \toprule
        \makecell{\textbf{Attn. Blocks}\\\textbf{Removed}} & \makecell{\textbf{Inference}\\ \textbf{Time Reduction}} \\
        \hline
        1 & 2.06\% \\
        4 & 8.24\% \\
        7 & 14.42\%  \\
        \bottomrule
    \end{tabular}
    \caption{Inference-time performance gains compared to blocks removed for {\small\texttt{Llama-3.1-8B-Instruct}}.}
    \label{tab:inference_gain}
\end{table}

It is also worth noting that when removing these attention blocks, we no longer have to store the corresponding parameters in memory, resulting in reduced-size models with lower GPU memory footprints. Each decoder layer in {\small\texttt{Llama-3.1-8B-Instruct}} consists of an attention block and an MLP block. The attention block includes four linear projections (query, key, value, and output), each with weight matrices of shape $h \times h$ (neglecting biases), contributing $4h^2$ parameters per layer. The MLP block typically includes two projections: $h \times 4h$ and $4h \times h$, totaling $8h^2$ parameters. Therefore, the attention block accounts for $\frac{4h^2}{4h^2 + 8h^2} = \frac{1}{3} \approx 33.3\%$ of the parameters in a decoder layer.

Given that {\small\texttt{Llama-3.1-8B-Instruct}} has 32 decoder layers and a total of $\sim$8B parameters, each layer contains approximately $250$M parameters. Removing one attention block saves roughly $83$M parameters, which is about $1.04\%$ of the model. Removing seven attention blocks yields a total reduction of $7 \times 83$M = $581$M parameters, or about $7.28\%$ of the total model size. Since each parameter occupies 2 bytes in float16, we can estimate the memory savings accordingly (Table \ref{tab:memory_savings}).

\begin{table}[h]
    \centering
    \begin{tabular}{ccc}
        \toprule
        \makecell{\textbf{Attn. Blocks}\\\textbf{Removed}} & \makecell{\textbf{Parameters}\\\textbf{Reduction}} & \makecell{\textbf{Memory}\\\textbf{Reduction}} \\
        \midrule
        1 & 1.04\% & $\sim$0.17 GB \\
        4 & 4.16\% & $\sim$0.67 GB \\
        7 & 7.28\% & $\sim$1.16 GB \\
        \bottomrule
    \end{tabular}
    \caption{Estimated memory savings by removing attention blocks from {\small\texttt{Llama-3.1-8B-Instruct}} (assuming 2 bytes per parameter in float16 thus total parameter memory of 16 GB).}
    \label{tab:memory_savings}
\end{table}

Further memory savings will be observed in practice due to reductions in optimizer states during training (e.g., Adam requires 8--12 bytes per parameter, leading to approximately 3$\times$ the model size), activation memory for forward and backward passes, gradient storage which is typically the same size as the model, and the KV cache used during inference with long contexts, which can consume 2--4 GB per 1,000 tokens for large models. This also means batch sizes can be potentially increased, further boosting inference gains.

\section{Conclusion}

We introduce Hopscotch, a simple yet effective method for skipping attention blocks in LLMs. To preserve model outputs, Hopscotch learns scaling factors for attention and MLP blocks in remaining layers, compensating for distributional changes in hidden states. We present promising results, achieving near-lossless performance on standard benchmarks. Hopscotch is also compatible with existing model compression techniques. We hope this work paves the way for future research on identifying and reducing redundant computations in attention mechanisms to build more efficient foundation models.

\section{Limitations}

We focus exclusively on the attention mechanism, including multi-head, multi-query, and group-query variants, which are widely adopted in open-source LLMs such as Llama and Qwen. However, emerging architectures like multi-head latent attention and state space models such as Mamba~\cite{gu2023mamba} offer new directions. It would be interesting to explore whether similar observations (e.g., skipping certain components) hold for these architectures as well. Additionally, there is an inherent trade-off between model compression and performance. When deploying compressed models in high-stakes applications that require high precision (such as disease detection) it is important to rigorously evaluate their performance to prevent potential harm.

\section{Ethical Considerations}

LLMs are growing which raises concerns about the accessibility and inclusiveness of AI development. Large models impose significant computational and memory demands, as well as increased infrastructure requirements. As a result, innovation increasingly concentrates within a few well-funded organizations, limiting participation from individual developers, smaller labs, and open-source community. This not only restricts opportunities for broader collaboration but also challenges the ability of open-source users to benefit from and contribute to cutting-edge model development. In this paper, we aim to bridge this gap by mitigating the inference cost of LLMs through architectural modifications, specifically, by identifying and skipping redundant attention blocks. Our hope is to contribute to a more equitable landscape in AI development, one where more individuals and institutions can meaningfully participate in and benefit from state-of-the-art LLMs.

\bibliography{references}

\clearpage
\appendix
\section{Appendix}

\subsection{Data and Loss Function Selection}
\label{app:loss}
Using GSM8K training data, we evaluated two different options for the ground-truth answers to be used in training. The first was to use the original ground-truth answers provided with the dataset. The second was to instead use the baseline model's greedy generations as the ground-truth answers, since the goal is to recover original model performance. To evaluate between the two options, we train with all scaling factors left free (no blocks removed) on both datasets. As we can see in Table \ref{tab:model_vs_data}, there seems to be a significant performance difference between the two options, with the model-generations-as-ground-truth scoring higher in both categories (flexible extract and strict match) for GSM8K. Additionally, we actually see improvement in strict match performance over the \textit{baseline} when doing model-generations-as-ground-truth training. We will see later as well that this holds even when blocks are removed from the model.

 \begin{table}[h]
     \centering
     \begin{tabular}{lcc}
         \toprule
         \textbf{Method} & \textbf{Flexible Extract} & \textbf{Strict Match} \\
         \hline
         Baseline & \textbf{83.10} & 66.94 \\
         Model GT & 82.79 & \textbf{77.48} \\
         Data GT & 60.88 & 42.61 \\
         \bottomrule
     \end{tabular}
     \caption{Comparison of different methods based on Flexible Extract and Strict Match metrics.}
     \label{tab:model_vs_data}
 \end{table}

This behavior makes sense given the goal of the training. When using the original data ground truth, we risk introducing \textit{new information} not previously in the model's distribution, and trying to learn that new information within a set of 128 total parameters could (and in this case did) lead to significant model degradation. Instead, what we are trying to accomplish is simply learning how to accent and distribute existing information to match expected steps, reasoning, formatting, etc. We are building a map from the existing attention blocks to the original known distribution, by learning from in-distribution outputs to find the appropriate attention block weighting. We are appropriately re-weighting the blocks in our model to approximate that original distribution, and so we need to ensure that we are learning exactly and only that reverse-mapping from output to block focus, rather than introducing anything the model was not originally capable of producing.

With the training data established, the next major piece to determine in the training dynamics was the actual loss function utilized. There were two options considered. The first was to use the standard Cross-Entropy loss typically employed in Causal LM training, in conjunction with our generated model-ground-truth answers. The second was to take things a step further, and instead use a loss calculated by the squared L2 norm between the final \textit{hidden states} (output of the final hidden layer) of the original model for a given input against the final hidden states of the model with removed attention blocks. This second method proved to be too rigid, however, as it's optimality relied on the assumption that the only path to a similarly correct final output was via a similar final hidden state.

\begin{figure}[t]
  \includegraphics[width=0.99\linewidth]{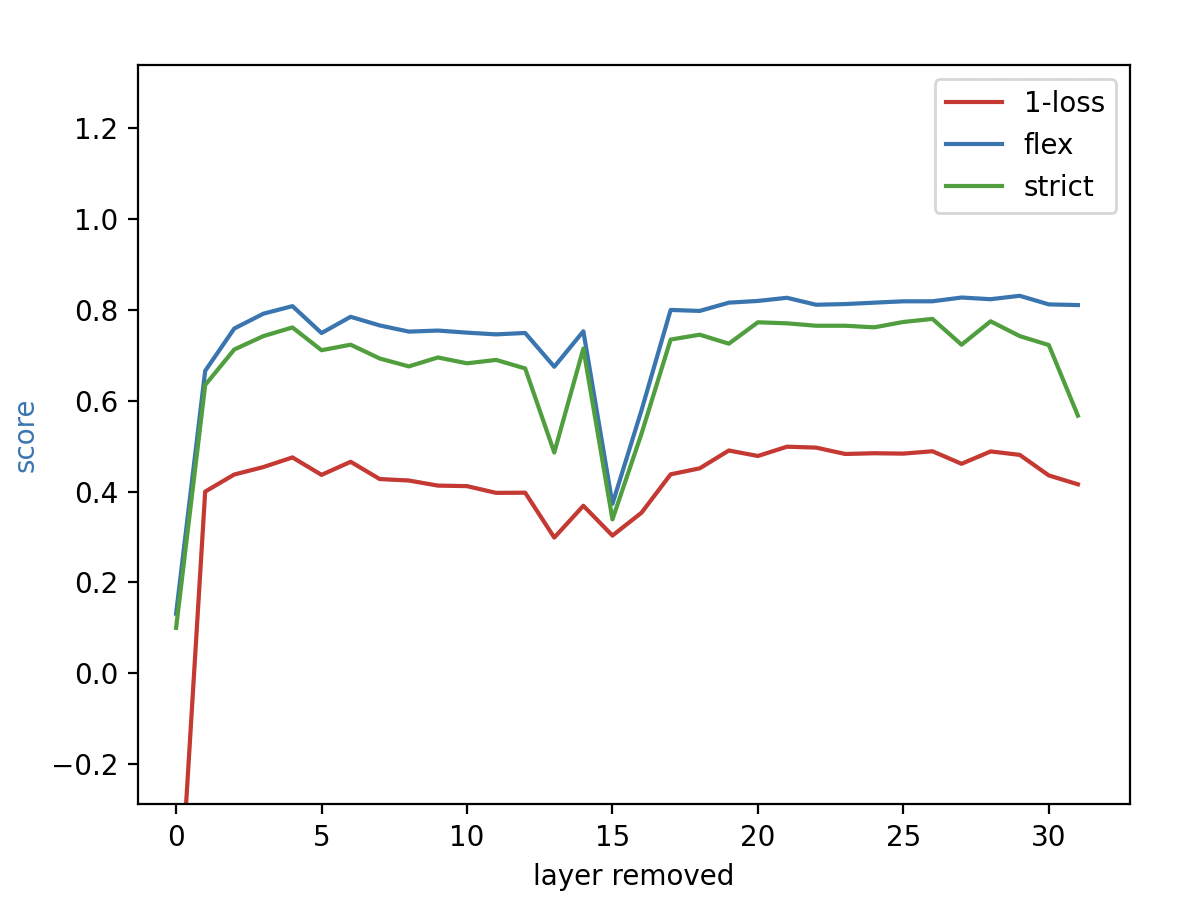} \hfill
  \caption {Correlation between average model-ground-truth cross-entropy loss given a layer with attention block removed and benchmark performance on GSM8K (flexible extract and strict match).}
  \label{fig:spearmann}
\end{figure}

Overall, the combination of model-outputs-as-ground-truth and Cross-Entropy loss resulted in promising initial results and high correlation to benchmark performance, with the training loss providing a Spearmann correlation of 0.9 for the GSM8K benchmark scores (Figure \ref{fig:spearmann}).

\subsection{Hyperparameters and Training Details}
\label{app:hyper}
Training is done using padding-free sample packing \cite{kundu2024enhancingtrainingefficiencyusing} with a batch size of 32, Adam optimizer \cite{kingma2017adammethodstochasticoptimization}, and a learning rate of 3e-3 for scaling parameter training, and 1e-2 for attention block selection. GSM8K training data consisted of 7473 samples. Evaluation done via LM Eval Harness \cite{eval-harness}.

\begin{figure*}[t]
\centering
  \includegraphics[width=0.48\linewidth]{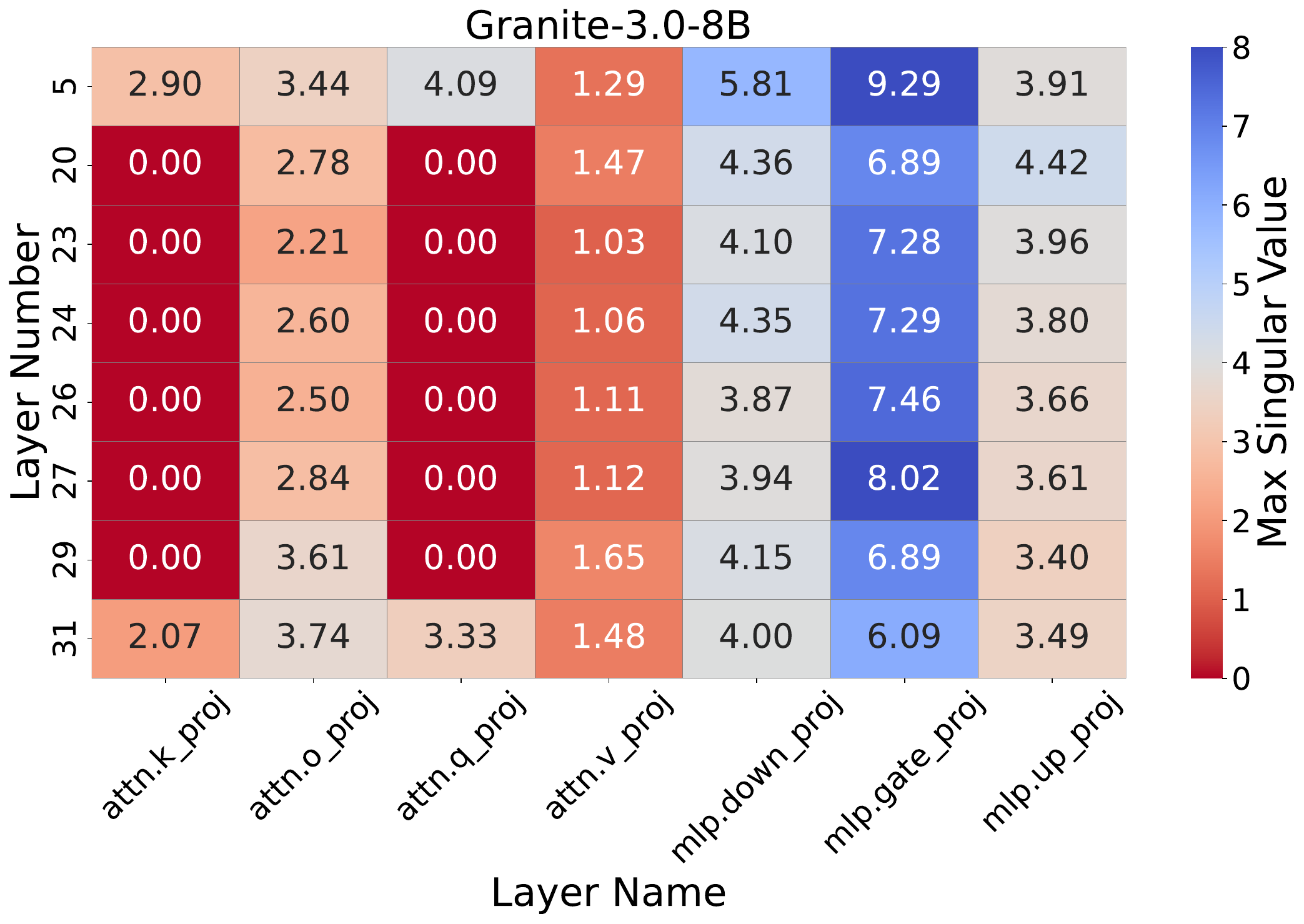} \hfill
  \includegraphics[width=0.48\linewidth]{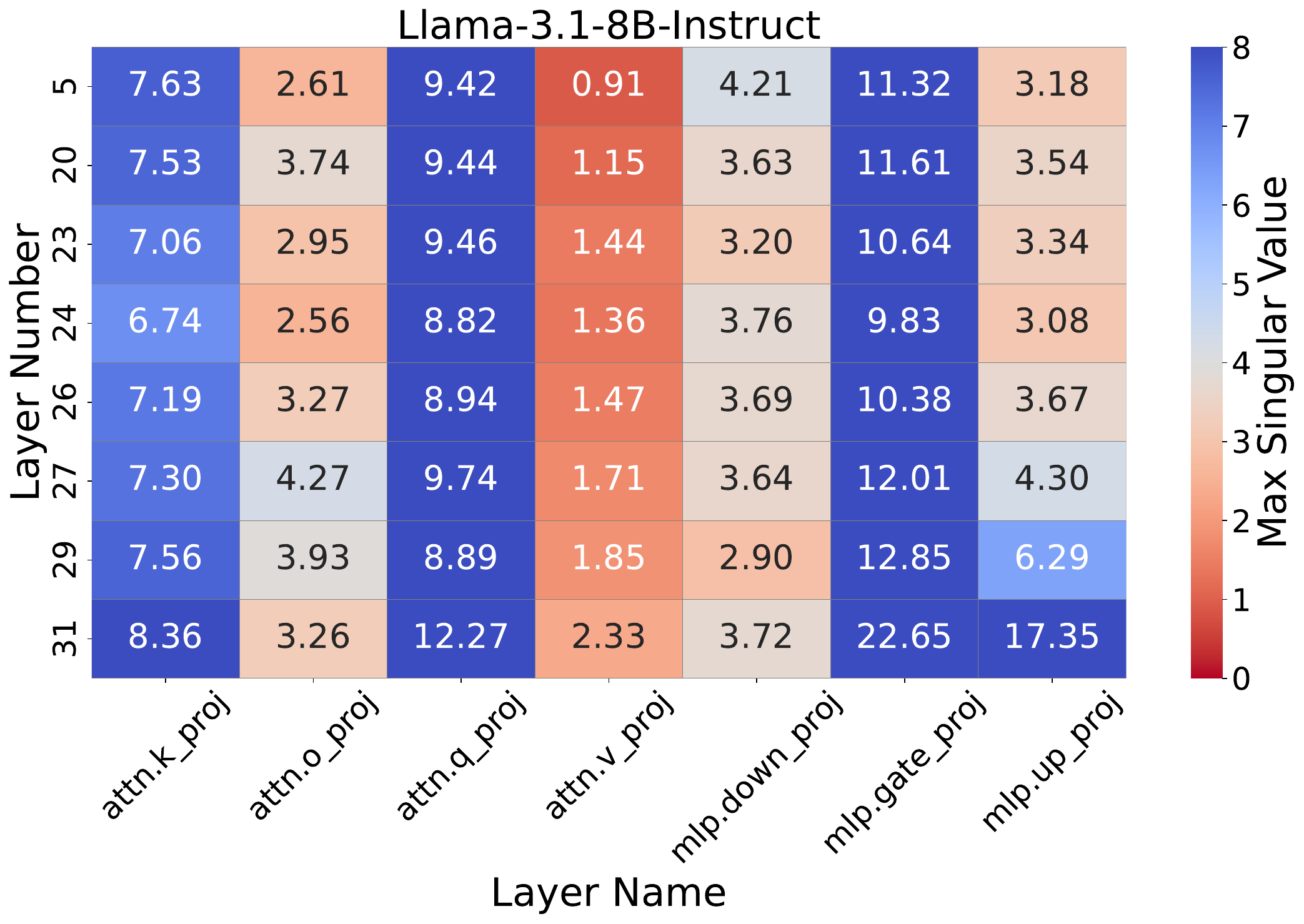}
  \caption {We show the largest singular values of weight matrices for a representative subset of layers to highlight structural differences between models. These values serve as a proxy for quantifying each block's contribution to the model's output. We present results using the \texttt{Granite-3.0-8B} model (left) and the \texttt{Llama-3.1-8B-instruct} model (right). Red indicates low singular values (near 0), while blue indicates high values (8 and above).}
  \label{fig:svd}
\end{figure*}

\subsection{Base Model Selection}
\label{app:model}

It is important to note that the selection of \texttt{Llama-3.1-8B-Instruct} as a base model was intentional, as it serves as an example of a model where all layers meaningfully transform the hidden states and contribute to the model's outputs, better showing the generalizability of Hopscotch. For example, we show that with a model like our fine-tuned \texttt{Granite-3.0-8B}, the process of selecting attention blocks to remove is trivial, as there are a number of non-contributing blocks in the model. We illustrate why it is possible to skip attention blocks by analyzing the largest singular values of the weight matrices across different layers. Singular values quantify how much a transformation can stretch or distort the input. Specifically, for a linear transformation $\bm{x} \to \mathbf{W} \bm{x}$, the inequality $\|\mathbf{W} \bm{x}\|_2 \leq \sigma_{\max}(\mathbf{W}) \|\bm{x}\|_2$ implies that if $\sigma_{\max}(\mathbf{W})$ is small and $\bm{x}$ is bounded, the output of the transformation will be near zero. This suggests that skipping such transformations would have minimal effect on the model's output.

We present results using two open-source LLMs: \texttt{Granite-3.0-8B} \citep{granite2024granite} and \texttt{Llama-3.1-8B-Instruct} \citep{grattafiori2024llama} in Figure~\ref{fig:svd}. The Granite model contains layers with negligible singular values, suggesting limited contribution to the output. \texttt{Llama-3.1-8B-Instruct} generally has larger and more consistent singular values. When we remove blocks from Llama, we see a non-negligible drop in benchmark performance. With Granite, however, we see in Table \ref{tab:granite_benchmarks} that removing attention blocks from layers with maximum singular values of zero results in no loss in model performance for the case of GSM8K.

\begin{table}[h]
    \centering
    \begin{tabular}{lccc}
        \toprule
        \multirow{1}{*}{\textbf{Bench}} 
        & \textbf{6 Blocks Skipped} & \textbf{Baseline}\\
        \hline
        gsm8k (flex.) & 75.66 & 74.00 \\
        gsm8k (strict) & 66.11 & 64.90 \\
        \bottomrule
    \end{tabular}
    \caption{Comparing benchmark performance when six attention blocks from layers (20, 23, 24, 26, 27, 29) are skipped, without any additional re-scaling. Includes original baseline scores before attention blocks were removed.}
    \label{tab:granite_benchmarks}
\end{table}

\subsection{Using Loss as an Approximation of Benchmark Performance}

As seen in Figure \ref{fig:spearmann}, training loss when removing blocks serves as strong indication of how the removal will affect overall model performance. Thus, at any given model state, we can know the best attention block to remove by finding the block that, when removed, provides the lowest loss. Since we simply need an indication via quick conversion, rather than an optimized loss, we can simply run our training method for a single epoch with an increased learning rate (1e-2), and take the average training loss as a comparable value for each block.

\subsection{Demonstrating Iterative Greedy Value}

\begin{table}[h]
    \centering
    \begin{tabular}{lcc}
        \toprule
        \textbf{Method} & \textbf{Average Loss (9 Blocks)} \\
        \hline
        Full Greedy & .2765 \\
        Iterative Greedy & \textbf{.2247} \\
        \bottomrule
    \end{tabular}
    \caption{Comparison of single-epoch average loss after nine attention blocks removed via full greedy approach and iterative greedy approach.}
    \label{tab:greedy_comp}
\end{table}

As can be seen in Table \ref{tab:greedy_comp}, when using the greedy iterative approach, the attention blocks removed prove to retain significantly more performance than taking a full greedy approach and assuming layer independence. With nine attention blocks removed from each, the average loss is notably lower for the greedy iterative approach. The iterative approach could remove two more attention blocks before reaching a similar loss to the full greedy approach.

\end{document}